\documentclass[hidelinks,10pt,conference]{IEEEtran}

\usepackage[utf8]{inputenc}
\usepackage[T1]{fontenc}

\usepackage{acronym}
\usepackage{cite}
\usepackage{dblfloatfix}
\usepackage{graphicx}
\usepackage{multirow}
\usepackage{xcolor}
\usepackage{amsfonts}

\usepackage{enumitem}

\usepackage{hyperref}
\usepackage[capitalise]{cleveref}

\setlist{nosep}

\crefname{paragraph}{Section}{Sections}

\title{Training Neural Networks with Internal State, Unconstrained Connectivity, and Discrete Activations}
\author{%
\IEEEauthorblockN{Alexander Grushin}%
    \IEEEauthorblockA{\small Galois, Inc.\\
        \texttt{agrushin@galois.com}}
}
\date{\today}

\usepackage[pscoord]{eso-pic}

\begin{document}
\bstctlcite{IEEEexample:BSTcontrol}

\maketitle
\thispagestyle{plain}
\pagestyle{plain}
\begin{abstract}
Today's most powerful machine learning approaches are typically designed to train stateless architectures with predefined layers and differentiable activation functions.  While these approaches have led to unprecedented successes in areas such as natural language processing and image recognition, the trained models are also susceptible to making mistakes that a human would not.  In this paper, we take the view that true intelligence may require the ability of a machine learning model to manage internal state, but that we have not yet discovered the most effective algorithms for training such models.  We further postulate that such algorithms might not necessarily be based on gradient descent over a deep architecture, but rather, might work best with an architecture that has discrete activations and few initial topological constraints (such as multiple predefined layers).  We present one attempt in our ongoing efforts to design such a training algorithm, applied to an architecture with binary activations and only a single matrix of weights, and show that it is able to form useful representations of natural language text, but is also limited in its ability to leverage large quantities of training data.  We then provide ideas for improving the algorithm and for designing other training algorithms for similar architectures.  Finally, we discuss potential benefits that could be gained if an effective training algorithm is found, and suggest experiments for evaluating whether these benefits exist in practice.
\end{abstract}

\section{Introduction} \label{section:intro}

Artificial intelligence has been witnessing major breakthroughs, with the release of very powerful {\it large language models} (LLMs), such as ChatGPT and GPT-4.  While these models can perform a wide range of highly complex tasks, ranging from question answering to software development, it is noteworthy that their underlying transformer-based architecture \cite{2017arXiv170603762V} is {\it stateless}, in that its output (e.g., prediction of the next token in a sentence) depends solely on the current input (previously-observed tokens, over some fixed-sized {\it context window}), without any memory of past inputs.  While this limitation can be mitigated by using a very large context window, this results in high computational overhead.  Perhaps more significantly, given that the biological brain maintains internal state/memory, a question arises: are there limitations on what a stateless model can achieve in practice, in terms of intelligence, when compared with a stateful model?  For example, is it possible that the limitations of current LLMs (e.g., ``hallucination'', i.e., the generation of false information \cite{koubaa2023gpt}, or the susceptibility to adversarial input perturbations \cite{zou2023}) can be, at least in part, attributed to a lack of state?

Unfortunately, it is presently very difficult to answer such questions, since the vast majority of existing stateful approaches, including recurrent neural networks (RNNs) such as the {\it long short-term memory} (LSTM) network \cite{HochSchm97}, have lagged far behind transformer-based architectures, in terms of performance.  However, this does not indicate that stateful approaches are less powerful than stateless ones; rather, it is possible that existing stateful models and their training methods (such as backpropagation through time \cite{robinson:utility}, which converts an RNN into a deep stateless network, with different layers representing activations at different time steps, and applies gradient descent) are not entirely appropriate for learning how to manage state, but that more effective techniques remain to be discovered.  Very recently, a study demonstrated a stateful model that outperformed transformers on multiple tasks \cite{gu2023}, and it will be interesting to observe whether this study may provide a turning point toward the use of state in machine learning.  The approach was based on an extension of {\it state-space models}, and used gradient-based training.  We have instead been exploring a different direction: training algorithms that do not rely on gradient descent, and can be applied to non-differentiable activation functions.

In this paper, we describe one attempt to find such a training algorithm for a stateful model, which consists of just two layers of neurons: the {\it input layer}, containing the current input $x_{t}$ and current state $h_{t}$, and the {\it hidden layer}, representing the state $h_{t+1}$ at the next time step; both inputs and states are represented as binary vectors.  A matrix $W$ of {\it weights}, along with the Heaviside step function, is used to map $x_{t}$ and $h_{t}$ to $h_{t+1}$, while the transpose $W^{T}$ of this matrix, also with the Heaviside step function, is used to reconstruct $x_{t}$ and $h_{t}$ from $h_{t+1}$; the reconstruction error is used as a training signal for adjusting the weights.  We train this model on a dataset of variable-length text samples obtained from news articles, and then use the model to convert each text sample into fixed-sized feature vector, where each element in the feature vector corresponds to the average activation of some hidden layer neuron over the duration of the sentence.  We find that the resulting feature vectors can then be used to classify samples (into topics) with an average accuracy of $82.2\%$, using linear regression.  The result suggests that it may potentially be feasible to develop effective training algorithms for architectures that are (a) stateful, (b) have very few initial constraints (we define just a single layer/matrix of weights $W$; it is up to the learning algorithm to potentially decompose it into sublayers/submatrices, if it so chooses), and (c) have discrete (specifically, binary) activations, which prohibit the use of gradient-based training techniques.

At the same time, we note that our accuracy is below those achieved by a number of other models, with some transformer-based approaches achieving an accuracy of over $95\%$ \cite{2019arXiv190505583S}; furthermore, we observe that our training algorithm is rather quick to converge, and does not benefit from large quantities of training data.  Later in this paper, we discuss possible ideas for further improving our training approach, as well as for uncovering other simple learning methods.  We also consider benefits that could potentially be achieved with an effective method for training stateful, unconstrained architectures with binary activations, and propose additional future work for evaluating whether (and to what extent) such benefits are realized.

\section{Related Work} \label{section:background}

How can a model learn to effectively manage its state and represent temporal sequences of inputs?  In developing our approach, we attempt to strike an appropriate balance between {\it stability} (i.e., old information is not forgotten, if it is relevant) and {\it adaptivity} (new information is effectively accommodated), while avoiding the storage of information that is unlikely to be needed in the future.  These conflicting objectives have been considered in past models; for example, in the aforementioned LSTM network \cite{HochSchm97}, specialized neurons called {\it input gates} and {\it forget gates} respectively determine the influence of a given input upon internal state (adaptivity), and the internal state’s persistence (stability); learning takes place through gradient descent.  These objectives are optimized in a very different way in the {\it linear autoencoder network} \cite{Sperduti2013LinearAN}, where the current input $x_{t}$ and state $h_{t}$ are linearly mapped to the next state $h_{t+1}$, such that $x_{t}$ and $h_{t}$ can be reconstructed as accurately as possible (also in a linear way) from $h_{t+1}$; stability is required to reconstruct $h_{t}$, while adaptability is required to reconstruct $x_{t}$.  Because the approach is linear, optimal parameters (weights) can be found via a closed-form expression; however, linearity also places significant limits on the functionality of the model.  Our approach is related to \cite{Sperduti2013LinearAN}, but introduces a nonlinearity (via the Heaviside step function) during both the generation of $h_{t+1}$ and the reconstruction of $x_{t}$ and $h_{t}$, and optimizes parameters in an approximate way.  Finally, we note that the two-layer architecture of our model is similar to that of a {\it restricted Boltzmann machine} (RBM) \cite{Smolensky1986}, and the reconstruction of $x_{t}$ and $h_{t}$ resembles RBM's reconstruction of visible activations from hidden activations; however, our model's activation dynamics are deterministic, and its learning procedure is different from that of an RBM \cite{carreiraperpinan2005contrastive}.

\section{Model} \label{section:model}

\begin{figure}[t]
    \centering
    \includegraphics[clip, trim={0in 3.00in 4.00in 1.5in}, width=.99\linewidth]
    {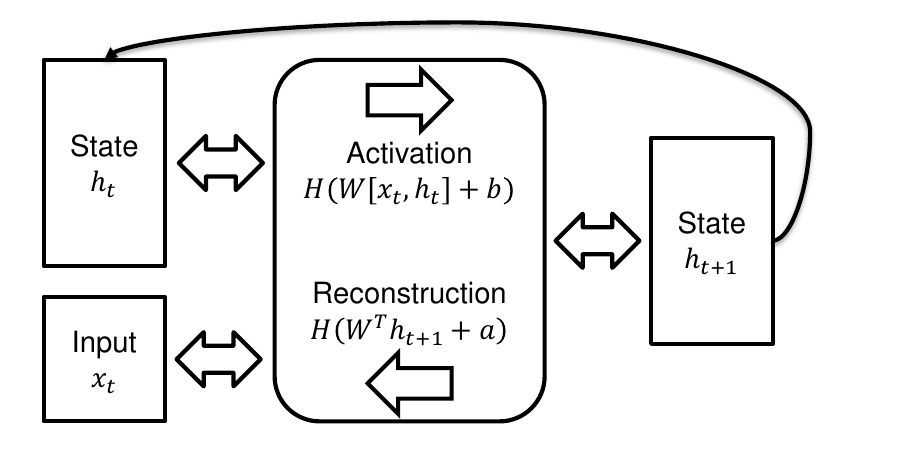}
    \caption{An illustration of our model, with equations for computing the activations within the hidden layer (with information flowing from the input layer on the left to the hidden layer on the right), and for reconstruction (with information flowing from right to left).  The thin black arrow denotes that the next state $h_{t+1}$, computed at time $t$, is provided to the input layer at time $t+1$, and can be viewed as a set of pairwise recurrent connections.}
    \label{fig:model}
\end{figure}

Our model is illustrated in \cref{fig:model}.  Formally, it consists of a matrix $W \in \mathbb{R}^{n \times (m+n)}$ of real-valued {\it weights}, a vector $a \in \mathbb{R}^{m+n}$ of {\it input biases}, and vector $b \in \mathbb{R}^{n}$ of {\it hidden biases}.  At time $0$, the {\it state} $h_{0}$ of the model is initialized to the zero vector of dimension $n$.  Given some binary input vector $x_{t} \in \{0, 1\}^{m}$ at time $t \geq 0$, the next state $h_{t+1} \in \{0, 1\}^{n}$ is computed as as follows:

\begin{equation} \label{eqn:activation}
h_{t+1} = H(W [x_{t}, h_{t}] + b)
\end{equation}

Here $H$ is the Heaviside step function (returning $1$ for any input greater than $0$, and $0$ for any other input), and $[x_{t}, h_{t}]$ is the concatenation of $x_{t}$ and $h_{t}$ (which is multiplied by $W$).

Conversely, given a state $h_{t+1}$, we may attempt to {\it reconstruct} the previous input $x_{t}$ and the previous state $h_{t}$ as follows:

\begin{equation} \label{eqn:reconstruction}
[x_{t}', h_{t}'] = H(W^{T} h_{t+1} + a)
\end{equation}

Here, $T$ is the transpose operator.  The {\it reconstruction error} is the difference $[x_{t}, h_{t}] - [x_{t}', h_{t}']$ between the true and the reconstructed input and state.  The model {\it learns} by attempting to reduce the reconstruction error, by adjusting the weights as follows:

\begin{equation} \label{eqn:learning:weights}
W \leftarrow W + h_{t} \otimes ( [r^{(x)}, r^{(h)}] \odot ( [x_{t}, h_{t}] - [x_{t}', h_{t}' ] ) )
\end{equation}

Here, $\otimes$ denotes the outer (tensor) product; $\odot$ denotes the Hadamard (element-wise) product; $r^{(x)}$ is the {\it input learning rate}, provided as a vector of dimension $m$ (where each of $m$ elements has the same value), and $r^{(h)}$ is the {\it state learning rate}, provided similarily as a vector of dimension $n$.  We analogously adjust the input biases as follows:

\begin{equation} \label{eqn:learning:biases:input}
a \leftarrow a + [r^{(x)}, r^{(h)}] \odot ( [x_{t}, h_{t}] - [x_{t}', h_{t}' ] )
\end{equation}

Finally, we adjust the hidden biases as follows:

\begin{equation} \label{eqn:learning:biases:hidden}
b \leftarrow b + r^{(h)} \odot (d - h_{t+1})
\end{equation}

Here, $d$ is the {\it density} parameter, provided as a vector of dimension $n$.  Whenever some some hidden neuron is inactive (i.e., the corresponding element of $h_{t+1}$ is $0$), the above rule increases the neuron's bias by $r^{(h)}d$; if it is active, then it decreases the bias by $r^{(h)}(1 - d)$; this attempts to achieve an average activation rate of $d$ for a given neuron (encouraging sparse representations if $d$ is low).

\section{Experiments} \label{section:experiments}

We train and evaluate our model on subsets of the AG's News Topic Classification Dataset\footnote{The preprocessed dataset, which we use, was obtained from \url{https://github.com/frederick0329/Text-Classification-Benchmark/tree/master/data/preprocessed}; it was derived from \url{http://groups.di.unipi.it/~gulli/AG_corpus_of_news_articles.html}.}.  The preprocessed dataset consists of short samples from news articles; each sample belongs to one of four topics: World, Sports, Business and Sci/Tech.  For any sample, we encode each character as a one-hot vector of $96$ elements, with each element corresponding to some letter, digit, or punctuation mark; for example, if the character is 'a', then the element corresponding to 'a' is set to $1$, with the rest of the elements set to $0$.  We train our model on $5000$ samples, selected at random from the AG's News training set, using the approach given in \cref{section:model}.  Each sample is provided to the model one character at a time; state $h_{t}$ is not reset when one sample ends and another begins; the $5000$ samples are provided only once.  We used the following parameters: $m = 96$ (given $96$ possible characters), $n = 4000$, $r^{(x)} = 0.01$, $r^{(h)} = 0.000001$, and $d = 0.1$.  Prior to training, each weight and bias was initialized by independently setting it to a random value drawn from a uniform distribution ranging between $-\frac{1}{m+n}$ and $\frac{1}{m+n}$.  We note that learning is unsupervised; i.e., topic labels are not used during the training process.

Once the model has been thus trained, we use it to convert an {\it additional} $5000$ samples (from the training set) into fixed-sized {\it feature vectors}.  For a given sample, with the first character occurring at time step $t^{(s)}$ and the last character occurring at time step $t^{(f)}$, the feature vector is computed as an average $\frac{1}{t^{(f)}-t^{(s)}+1} \sum_{t=t^{(s)}}^{t^{(f)}} h_{t+1}$ of states $h_{t+1}$ obtained after each character has been presented; each element of the feature vector can be viewed as the proportion of times that some hidden neuron is active, i.e., has a value of $1$.  Again, state is not reset between samples; also, weights and biases are not modified during this process.  For each of the $5000$ samples, we thus obtain a feature vector of $n = 4000$ elements, and we then use ridge regression to train a classifier on these feature vectors, with topic labels provided as ground truth.  Finally, we generate feature vectors from the $7600$ samples in the AG's News test set (using the same procedure as described above), apply the trained classifier to them, and measure classification accuracy.  This process is repeated over $5$ trials, each time using a different set of random initial values for weights and biases, and a different, randomly-chosen subset of $10,000$ samples from AG's News training set ($5000$ for unsupervised model training and $5000$ for supervised classifier training); training subsets are disjoint across the $5$ trials, but the same test set is used in each trial.  The classification accuracy for each trial is reported in \cref{table:results}, in the ``Model'' row; we note that the choice of initial parameter values and training subset has only a minor effect on accuracy.

\begin{table}[]
\centering
\caption{Experimental Results}
\label{table:results}
{\footnotesize
\begin{tabular}{|c|c|c|c|c|c|}
\hline
Trial & 1      & 2      & 3      & 4      & 5      \\ \hline
Model     & 82.3\% & 82.6\% & 81.9\% & 82.1\% & 82.2\% \\ \hline
Baseline     & 49.9\% & 49.2\% & 48.9\% & 48.4\% & 49.4\% \\ \hline
\end{tabular}

}
\end{table}

As a baseline for comparison, we reran the experiment without our model, but rather, with feature vectors obtained as averages $\frac{1}{t^{(f)}-t^{(s)}+1} \sum_{t=t^{(s)}}^{t^{(f)}} x_{t}$ of {\it inputs}, i.e., as vectors of $96$ elements, where each element's value is the frequency of occurrence of a given character in the sample.  Interestingly, even with this very basic strategy, classification performance is well above the $25\%$ that one would expect with random guessing (given each of four topics occurs in the test set the same number of times), as we indicate in the ``Baseline'' row of \cref{table:results}.  Still, its performance is significantly below that of our model, which suggests that the model does not merely store individual characters as part of its state, but computes more complex (and useful) features.

\begin{figure}[t]
    \centering
    \includegraphics[clip, width=.99\linewidth]
    {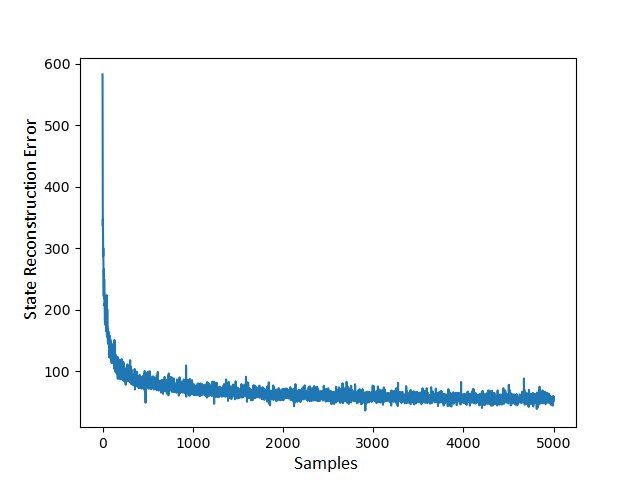}
    \caption{State reconstruction error during unsupervised training, for Trial 1.}
    \label{fig:error}
\end{figure}

It is noteworthy that our result is achieved while using only a small subset of the available training set (which consists of $120,000$ samples) for training.  While the ability to learn from limited data can be viewed as a positive feature, there is also a significant limitation, in that the model does not appear to benefit significantly from larger training datasets.  As we show in \cref{fig:error}, the reconstruction error, here defined as the Hamming distance between $h_{t}$ and $h_{t}'$, initially decreases rapidly, but soon becomes relatively flat.  (The input reconstruction error, which we do not display, is typically $0$, with occasional small spikes, except at the very beginning of the training process.)  Thus, simply using a larger training set is unlikely to fill the accuracy gap between our approach and state of the art approaches such as \cite{2019arXiv190505583S}, and further modifications are needed.

\section{Discussion and Future Work} \label{section:future}

Given the above limitation, a short-term goal is to determine whether there are effective ways to slow down the convergence of our learning procedure, such that it makes use of more training examples to find a more optimal set of weights and biases.  At the same time, it is important to note that lower reconstruction error alone does not imply a better representation; for example, if the model simply learns to compute $h_{t+1}$ as the zero vector, and to reconstruct both $x_{t}$ and $h_{t}$ as zero vectors, then input reconstruction error will be $1$ (given that the input is a one-hot vector), while memory reconstruction error will be $0$, with the overall reconstruction error being much lower than what we observed in our experiments, but with an ultimately useless representation.  Both the procedure for learning hidden biases (\cref{eqn:learning:biases:hidden}) and the large difference between the input and state learning rates (\cref{section:experiments}) prevent this from happening.  One possible approach for improving the training algorithm is to not only use a distinct learning rate for the input, but to also associate different learning rates with different elements of the reconstructed state vector $h_{t}'$, and to adapt these learning rates automatically in some way.  Other possibilities include introducing a controlled degree of randomness into the activations and/or the weights/biases, in order to slow down convergence, or changing the weight and bias initialization scheme, e.g., with most weights and biases being initially assigned a value of $0$.

It is also of interest to explore alternative learning rules for similar stateful, unconstrained architectures with discrete activations.  One possible direction is to explore the space of purely {\it local} rules, where the weight change $\Delta w_{ji}$ on the connection between neurons $i$ and $j$ depends only upon the (binary) activations of $i$ and $j$ at times $t$ and $t+1$ (or, alternatively, $t+1$ and $t+2$ in the case of $j$), without requiring any additional computation (such as the reconstruction of the previous state and input in our current approach).  Such a rule can be represented as a table of $2^{4} = 16$ rows (given that two neurons at two time steps give four activation values, where each activation value can be either $0$ or $1$); for each row, we can specify whether the weight should increase, decrease, or stay the same.  If we further enforce that whenever a weight is increased or decreased, it is always by the same magnitude, then the number of possible rules is $3^{16}=43,046,721$.  While this rule space is large, it is not intractable to explore; some rules can be filtered out after rapidly failing on very simple learning problems with very small neural network sizes, with successful rules then being applied to more complex problems with larger networks.  This exploration is somewhat analogous to (and inspired by) the characterization of the much smaller space of $256$ rules for elementary cellular automata \cite{citeulike:106131}; the exploration led to the discovery of a very simple rule that is capable of supporting universal computation.

If a training approach (whether an extension of the one presented in \cref{section:model} or a different one) is sufficiently successful at learning useful hidden representations from large temporal datasets, then it would be of interest to investigate whether the representations provide benefits other than just improved classification accuracy.  Such benefits might potentially include the following:

\begin{itemize}
    \item {\it The ability to remember information from the distant past:} A theoretical benefit of a stateful model is that it can retain even old information, if it is relevant, though experiments are needed to determine whether this is the case in practice.  If so, then in principle, a stateful model could be used as a preprocessor to predictive models, such as LLMs: rather than providing information (e.g., text tokens) over a fixed-sized context window as input to an LLM, we could provide the state(s) of the stateful model\footnote{Here, it is of interest to note that while an LLM learns to predict the next input, our stateful model learns to recall the previous input and previous state.}.  However, further research is needed to determine how many states to provide, and in what form.  In our experiments, we found that if we simply use the state $h_{t^{(f)}+1}$ after the end of the sentence as a feature vector, classification accuracy is much lower than if we use the average state over the entire sentence $\frac{1}{t^{(f)}-t^{(s)}+1} \sum_{t=t^{(s)}}^{t^{(f)}} h_{t+1}$; it is not clear whether such an average would serve as an appropriate input, whether some other postprocessing should be applied to states, or whether some set of recent states should be provided as input to an LLM in their raw form.  Another possibility is to modify the existing architecture such that it not only computes the next state $h_{t+1}$, but also predicts the next input $x_{t+1}$; i.e., to include prediction as an additional task to be performed by the model.
    \item {\it Robustness to adversarial perturbations:} It is well-known that current machine learning models can be susceptible to adversarial attacks, where small changes in the input, considered insignificant (or even imperceptible) by a human, can result in large (and incorrect) output changes \cite{2013arXiv1312.6199S,zou2023}.  As a dynamical system, a stateful model could potentially learn to stabilize the trajectory, and make it more robust to perturbations in the input; further robustness might perhaps be afforded by the use of binary, rather than real-valued activations.  To determine whether this is the case, experiments can be performed with gradient-free attacks (e.g. \cite{2018arXiv180205666U}), gauging their effect on classification accuracy.
    \item {\it Improved explainability/interpretability:} It is of interest to determine whether the approach allows for more insight into how the model makes its decisions, compared with conventional approaches.  For the model presented in \cref{section:model}, we can apply \cref{eqn:reconstruction} multiple times to attempt to reconstruct not only the previous input $x_{t}$ and state $h_{t}$, respectively, but inputs and states at earlier time steps as well.  By determining which past inputs and state elements (features) are reconstructed more or less accurately, it may be possible to gauge the extent to which they are used in decision-making; the process might be simplified by the binary (rather than real-valued) nature of the inputs and states.  An interesting research question is whether reconstruction errors provide more insight than gate values within an LSTM network.
    \item {\it Improved verifiability:} Another relevant question is whether the binary inputs and states can make the model more amenable to the application of formal methods, in order to prove certain properties about the model.  For example, there are techniques for converting a recurrent neural network into a finite state machine, but they require the discretization of the real-valued inputs and states \cite{koul2018}, which can be non-trivial; this discretization would not be necessary for the model given here.
    \item {\it Low power consumption:} The use of binary activations, along with the gradient-free training approach, may potentially allow the model to be implemented on low-power hardware.  However, experiments are necessary to determine the extent to which the precision of weights and biases can be reduced, without significantly affecting the effectiveness of training or the operation of a trained model.
    \item {\it Ease of design and customization:} Rather than designing a modular architecture with prespecified layers, our approach has just a single layer of weights, which reduces the number of tunable hyperparameters (such as the size of each layer and the number of layers).  If the training of such an architecture is effective for a wide range of problems, then less customization is required, given a new problem, but to determine whether this is the case, experiments must be performed with a diverse range of problems.  It is also of interest to analyze the trained models, and determine whether modularity emerges as a result of training, where the weight matrix $W$ is decomposed into sublayers/submatrices, with near-zero weights assigned to connections between sublayers.
    
\end{itemize}

It is, of course, not guaranteed that these benefits will be realized; nonetheless, it is the author's belief that it is worth continuing to explore alternative approaches to training stateful neural networks, as a promising direction in the quest for artificial general intelligence.

\section*{Acknowledgments}

The author wishes to thank Galois, Inc. for supporting this effort, and in particular, Walt Woods, Shpat Morina, and Adam Karvonen for useful discussions and advice.

\newcommand\lastcolumnfix{\enlargethispage{-11cm}}
\IEEEtriggercmd{\lastcolumnfix}
\bibliographystyle{sty/ieee/IEEEtran-nomonth}
\bibliography{sty/ieee/IEEEabrv,references}

\begin{thebibliography}{10}
\def\url#1{}
\csname url@samestyle\endcsname
\providecommand{\newblock}{\relax}
\providecommand{\bibinfo}[2]{#2}
\providecommand{\BIBentrySTDinterwordspacing}{\spaceskip=0pt\relax}
\providecommand{\BIBentryALTinterwordstretchfactor}{4}
\providecommand{\BIBentryALTinterwordspacing}{\spaceskip=\fontdimen2\font plus
\BIBentryALTinterwordstretchfactor\fontdimen3\font minus
  \fontdimen4\font\relax}
\providecommand{\BIBforeignlanguage}[2]{{%
\expandafter\ifx\csname l@#1\endcsname\relax
\typeout{** WARNING: IEEEtran.bst: No hyphenation pattern has been}%
\typeout{** loaded for the language `#1'. Using the pattern for}%
\typeout{** the default language instead.}%
\else
\language=\csname l@#1\endcsname
\fi
#2}}
\providecommand{\BIBdecl}{\relax}
\BIBdecl

\bibitem{2017arXiv170603762V}
A.~{Vaswani} \emph{et~al.}, ``{Attention Is All You Need},'' \emph{arXiv
  e-prints arXiv:1706.03762}, 2017.

\bibitem{koubaa2023gpt}
A.~Koubaa, ``Gpt-4 vs. gpt-3.5: A concise showdown,'' 2023.

\bibitem{zou2023}
A.~{Zou} \emph{et~al.}, ``{Universal and Transferable Adversarial Attacks on
  Aligned Language Models},'' \emph{arXiv e-prints arXiv:2307.15043}, 2023.

\bibitem{HochSchm97}
S.~Hochreiter \emph{et~al.}, ``Long short-term memory,'' \emph{Neural
  Computation}, vol.~9, no.~8, pp. 1735--1780, 1997.

\bibitem{robinson:utility}
A.~J. Robinson \emph{et~al.}, ``The utility driven dynamic error propagation
  network,'' Engineering Department, Cambridge University, Cambridge, UK, Tech.
  Rep. CUED/F-INFENG/TR.1, 1987.

\bibitem{gu2023}
A.~{Gu} \emph{et~al.}, ``{Mamba: Linear-Time Sequence Modeling with Selective
  State Spaces},'' \emph{arXiv e-prints arXiv:2312.00752}, 2023.

\bibitem{2019arXiv190505583S}
C.~{Sun} \emph{et~al.}, ``{How to Fine-Tune BERT for Text Classification?}''
  \emph{arXiv e-prints arXiv:1905.05583}, 2019.

\bibitem{Sperduti2013LinearAN}
A.~Sperduti, ``Linear autoencoder networks for structured data,'' in
  \emph{Ninth International Workshop on Neural-Symbolic Learning and
  Reasoning}, 2013.

\bibitem{Smolensky1986}
P.~Smolensky, ``Information processing in dynamical systems: Foundations of
  harmony theory,'' in \emph{Parallel distributed processing: Explorations in
  the microstructure of cognition}.\hskip 1em plus 0.5em minus 0.4em\relax
  Cambridge, MA: MIT Press, 1986, pp. 194--281.

\bibitem{carreiraperpinan2005contrastive}
M.~A. Carreira-Perpinan \emph{et~al.}, ``On contrastive divergence learning,''
  A.~Intelligence \emph{et~al.}, Eds., 2005.

\bibitem{citeulike:106131}
\BIBentryALTinterwordspacing
S.~Wolfram, \emph{A New Kind of Science}.\hskip 1em plus 0.5em minus
  0.4em\relax {Wolfram Media}, 2002.
  \url{http://www.amazon.com/exec/obidos/redirect?tag=citeulike07-20\&path=ASIN/1579550088}
\BIBentrySTDinterwordspacing

\bibitem{2013arXiv1312.6199S}
C.~{Szegedy} \emph{et~al.}, ``{Intriguing properties of neural networks},''
  \emph{arXiv e-prints arXiv:1312.6199}, 2013.

\bibitem{2018arXiv180205666U}
J.~{Uesato} \emph{et~al.}, ``{Adversarial Risk and the Dangers of Evaluating
  Against Weak Attacks},'' \emph{arXiv e-prints arXiv:1802.05666}, 2018.

\bibitem{koul2018}
A.~{Koul} \emph{et~al.}, ``{Learning Finite State Representations of Recurrent
  Policy Networks},'' \emph{arXiv e-prints arXiv:1811.12530}, 2018.

\end{thebibliography}


\end{document}